\documentclass[letterpaper]{article}

\usepackage{aaai2026}
\usepackage{times}
\usepackage{helvet}
\usepackage{courier}
\usepackage[hyphens]{url}
\usepackage{graphicx}
\usepackage{natbib}
\usepackage{caption}
\usepackage{multirow}
\usepackage{amsmath,amssymb,amsfonts}
\usepackage{booktabs}
\usepackage{subcaption}

\let\aaaiorigbibliographystyle\bibliographystyle
\renewcommand{\bibliographystyle}[1]{}

\AtBeginDocument{\setcitestyle{numbers,square,sort&compress}}
\makeatletter
\def\subsubsection{\@startsection{subsubsection}{3}{\z@}{-6pt plus -2pt minus -1pt}{-1em}{\normalsize\bf}}
\makeatother

\newcommand{\vs}{\textit{vs.}}
\newcommand{\eg}{\textit{e.g.,}}

\newcommand{\seclabel}[1]{\label{sec:#1}}
\newcommand{\secref}[1]{Section~\ref{sec:#1}}
\newcommand{\figlabel}[1]{\label{fig:#1}}
\newcommand{\figref}[1]{Fig.~\ref{fig:#1}}
\newcommand{\tablabel}[1]{\label{tab:#1}}
\newcommand{\tabref}[1]{Table~\ref{tab:#1}}
\newcommand{\arxiv}{arXiv}
\newcommand{\latex}{\LaTeX{}}

\title{AI-Assisted Peer Review at Scale: The AAAI-26 AI Review Pilot}

\author{
Joydeep Biswas\textsuperscript{\rm 1},
Sheila Schoepp\textsuperscript{\rm 2},
Gautham Vasan\textsuperscript{\rm 2},
Anthony Opipari\textsuperscript{\rm 3},
Arthur Zhang\textsuperscript{\rm 1},
Zichao Hu\textsuperscript{\rm 1},
Sebastian Joseph\textsuperscript{\rm 1},
Matthew Lease\textsuperscript{\rm 4},
Junyi Jessy Li\textsuperscript{\rm 5},
Peter Stone\textsuperscript{\rm 1,7},
Kiri L. Wagstaff\textsuperscript{\rm 6},
Matthew E. Taylor\textsuperscript{\rm 2,8},
Odest Chadwicke Jenkins\textsuperscript{\rm 3}
}

\affiliations{
\textsuperscript{\rm 1}Department of Computer Science, The University of Texas at Austin;
\textsuperscript{\rm 2}Department of Computing Science, University of Alberta;
\textsuperscript{\rm 3}Department of Electrical Engineering and Computer Science, University of Michigan;
\textsuperscript{\rm 4}School of Information, The University of Texas at Austin;
\textsuperscript{\rm 5}Department of Linguistics, The University of Texas at Austin;
\textsuperscript{\rm 6}The Valley Library, Oregon State University;
\textsuperscript{\rm 7}Sony AI;
\textsuperscript{\rm 8}Alberta Machine Intelligence Institute.\\
joydeepb@cs.utexas.edu, sschoepp@ualberta.ca, vasan@ualberta.ca, topipari@umich.edu, arthurz@utexas.edu, zichao@utexas.edu, sebaj@utexas.edu, ml@utexas.edu, jessy@utexas.edu, pstone@cs.utexas.edu, kiri.wagstaff@oregonstate.edu, matthew.e.taylor@ualberta.ca, ocj@umich.edu
}

\begin{document}

\maketitle

\begin{abstract}

Scientific peer review faces mounting strain as submission volumes surge, making it increasingly difficult to sustain review quality, consistency, and timeliness.
Recent advances in AI have led the community to consider its use in peer review,
yet a key unresolved question is whether AI can generate technically sound reviews at real-world conference scale.
Here we report the first large-scale field deployment of AI-assisted peer review: every main-track submission at AAAI-26 received one clearly identified AI review from a state-of-the-art system.
The system combined frontier models, tool use, and safeguards in a multi-stage process to generate reviews for all 22,977
full-review papers in less than a day.
A large-scale survey of AAAI-26 authors and program committee members showed that participants not only found AI reviews useful, but actually preferred them to human reviews on key dimensions such as technical accuracy and research suggestions.
We also introduce a novel benchmark and find
that our system substantially outperforms a simple LLM-generated review baseline at detecting a variety of scientific weaknesses.
Together, these results show that state-of-the-art AI methods can already make meaningful contributions to scientific peer review at conference scale, opening a path toward the next generation of synergistic human-AI teaming for evaluating research.

\end{abstract}

\section{Introduction}
The scientific peer review process is under significant strain.  The AAAI Conference on Artificial Intelligence, a major artificial
intelligence (AI) research conference, received more than 30,000 initial
submissions for 2026\footnote{The AAAI-26 review process ran from Aug.--Nov.\ 2025.}, up from approximately 15,000 for 2025. This dramatic growth
is not unique to AAAI; submissions have
grown rapidly for other venues too, such as Nature~\cite{natureindex2025growth} and NeurIPS ~\cite{tanmirage}.

Unfortunately, despite this rapid growth, the peer review
process has remained largely static, with a large cohort of human reviewers providing detailed reviews and ratings for papers, and a smaller group of senior
researchers comparing those reviews and ratings to make final paper acceptance
recommendations.

While this established review process has long endured, the rising scale of submissions means we face increasingly overburdened reviewers with more papers assigned, the need to recruit an ever-wider pool of potentially less experienced reviewers, and increasingly compressed timelines.

Maintaining the quality, consistency, and
timeliness of peer review is thus
increasingly challenging. For example, the scale of AAAI-26 submissions required the recruitment and oversight of over 28,000 Program Committee members, Senior Program Committee members, and Area Chairs, nearly three times the size of the committee in AAAI-25~\cite{aaai26_review_process_update}.

At the same time, there have been rapid advances in state-of-the-art AI
systems, particularly in mathematical, coding~\cite{deepseek-math,deepseek-code}, and other technical domains~\cite{gpqa}.

Autonomous AI scientists~\cite{aiscientist,jansen-etal-2025-codescientist,si2026towards,garikaparthi2026researchgym,tanmirage} now perform iterative feedback loops of automated paper writing, generating critiques, and revising the writing in response. 

Most relevant, a

growing body of work is now
investigating whether and how AI systems can be used to assist with
scientific peer review and help alleviate the growing strain on the review
process~\cite{du-etal-2024-llms,liang2024usefulfeedback,liang2024monitoring,liu2023reviewergpt,yu-etal-2024-automated,zhou-etal-2024-llm,bougie2025gar,gao2025reviewagents,shin-etal-2025-mind,xu-etal-2025-llms-identify}.
Against the backdrop of increasing strain on human peer review, the reviewer population has started using AI reviewing against explicit guidance~\cite{naddaf2026more}, 
while conference venues grapple with the question of how to effectively and meaningfully integrate AI into the review process in a way that is beneficial to the community.

There have been synthetic, benchmark-based, and post-hoc studies of
AI-generated reviews and AI-assisted peer review on existing papers and
reviews, including retrospective analyses of reviewers' use of AI during peer
review~\cite{liang2024usefulfeedback,liu2023reviewergpt,zhou-etal-2024-llm,gao2025reviewagents,bougie2025gar,liang2024monitoring,latona2024aireviewlottery}.
Evaluation remains challenging, however, because existing benchmarks and
evaluation datasets measure only limited aspects of reviewing. These aspects
include specific error types, evaluation of structured outputs as opposed to
unstructured review text, or similarity of scores to human reviews, rather than
end-to-end review quality~\cite{xi2025flaws,sahu2025reviewertoo,lou2025aaar,garg-etal-2025-revieweval}.

Based on these encouraging findings, there have been a small number of live
studies of AI systems providing limited assistance within conference
workflows, notably 
author checklist assistance in NeurIPS~2024~\cite{goldberg2024checklistassistant}
and feedback to reviewers in ICLR~2025~\cite{thakkar2025iclrfeedback}.
However, neither of these studies deployed official AI-generated reviews on
live submissions. Prior to AAAI-26, there had been no conference-wide live
study of AI-generated reviews deployed on real submissions at a major
conference. Thus, despite substantial recent progress, a key question remained: \emph{could state-of-the-art AI systems generate technically meaningful
and practically useful reviews in a live peer-review process at conference
scale?}

The AAAI-26 AI Review Pilot Program was the first
full-scale live study of AI-generated reviews on real
submissions at a major conference. Every paper that entered the full review phase (22,977 in total)
in the main track at AAAI-26 received one clearly labeled AI review,
generated by a state-of-the-art custom-developed AI review system. 
Consistent with prior results, we found that simply asking off-the-shelf LLMs to review papers does \emph{not} lead to high-quality reviews~\cite{liu2023reviewergpt,zhou-etal-2024-llm,zhang2025reviewingscientificpapers}. Recent work has therefore explored more structured review systems based on deeper multi-stage reasoning, hierarchical question decomposition, and multimodal  workflows~\cite{zhu2025deepreview,chang2025treereview,lu2025agentreviewers}. 
In response, we developed a novel, multi-stage, multi-tool, LLM-based review pipeline that \emph{does} lead to very high-quality reviews.
AAAI-26 used a double-blind review process, so reviewers and authors were anonymized to one another during evaluation, but both reviewers and authors could identify the AI review.
The AI
review was added during Phase 1
of the two-phase review process, alongside at least two human
reviews. The AI review system did not include any scores or recommendations,
and no human reviewers were replaced in the process. Instead, the AI reviews
were intended to provide additional input to the peer-review process \cite{renata2025ai}.

Senior
Program Committee members (SPCs) and Area Chairs (ACs) (who are responsible for making paper recommendations and normalizing within their batch),
were able to view the AI
reviews along with the human reviews and use them to help make their decisions about whether
to promote papers to Phase 2 of the review process. Papers that were promoted
to Phase 2 received additional human reviews, and the authors had a chance to
respond to all reviews, including the AI review, before the reviewers, SPCs,
and ACs discussed the papers and made their final decisions in light of all reviews, including the AI review. 

An optional survey was sent to authors, reviewers, SPCs, and ACs to assess both
human and AI reviews on a variety of criteria. The design, development, and
deployment of the AI Review Pilot Program was overseen through extensive and
careful consideration by the AAAI Executive Council, the AAAI Conference
Committee, the AAAI Ethics Committee, and the AAAI-26 Program Committee. The
survey design was reviewed by the Institutional Review Boards (IRB) of 
the University of Texas at Austin and the University of Michigan, as well as the Research Ethics Board (REB) of the University of Alberta.

We found that AI-generated peer reviews are operationally feasible at conference
scale, with a modest cost of less than \$1 per paper
(covered through an in-kind donation of API credits from OpenAI as a AAAI-26 sponsor)\footnote{Even with 30K submissions, this cost is a small fraction of the conference budget.}. All reviews were generated
in less than 24 hours using a state-of-the-art frontier LLM in a
multi-stage workflow with  coding and web search tools. 
By creating a new
benchmark for scientific review based on synthetic perturbations to published
papers, we further showed that the AAAI-26 AI Review System significantly improves upon the
ability of a base LLM to catch scientific errors in the story, presentation,
evaluations, correctness, and significance of scientific papers. 

Finally, analysis of the
survey of authors, reviewers, SPCs, and ACs (5,834 responses)
indicates that the community not only found the AAAI-26 AI reviews helpful, but that AI
reviews were actually preferred to human reviews on several important criteria spanning
technical accuracy, review focus, and research suggestions. Moreover, 
respondents believed that AI reviews would be useful in future peer review
processes. 

Qualitative responses further highlighted both the success and the
current limitations of the system. Strengths mentioned by respondents included providing an
impartial review as a safeguard against human variability and some forms of malicious activity; and structured and
in-depth technical feedback.
Weaknesses included errors in reading some
equations and tables, difficulty in prioritizing the significance of issues (an area of ongoing research),
and producing reviews that were longer than readers preferred (a limitation that is straightforward to mitigate via tighter output-length controls).

Overall, we found that the technical capabilities of this AI review system are
already sufficient to usefully assist in scientific peer review in ways that the
community finds helpful. Further study is needed to ascertain how best to
integrate the complementary strengths of AI systems and human reviewers to
improve the process of evaluating and advancing scientific research.

\section{The AAAI-26 AI Review System}
\begin{figure*}[htb]
    \centering
    \begin{subfigure}[c]{0.54\textwidth}
        \centering
        \includegraphics[width=\textwidth, trim=3.25cm 1.125cm 3.25cm 1.125cm, clip]{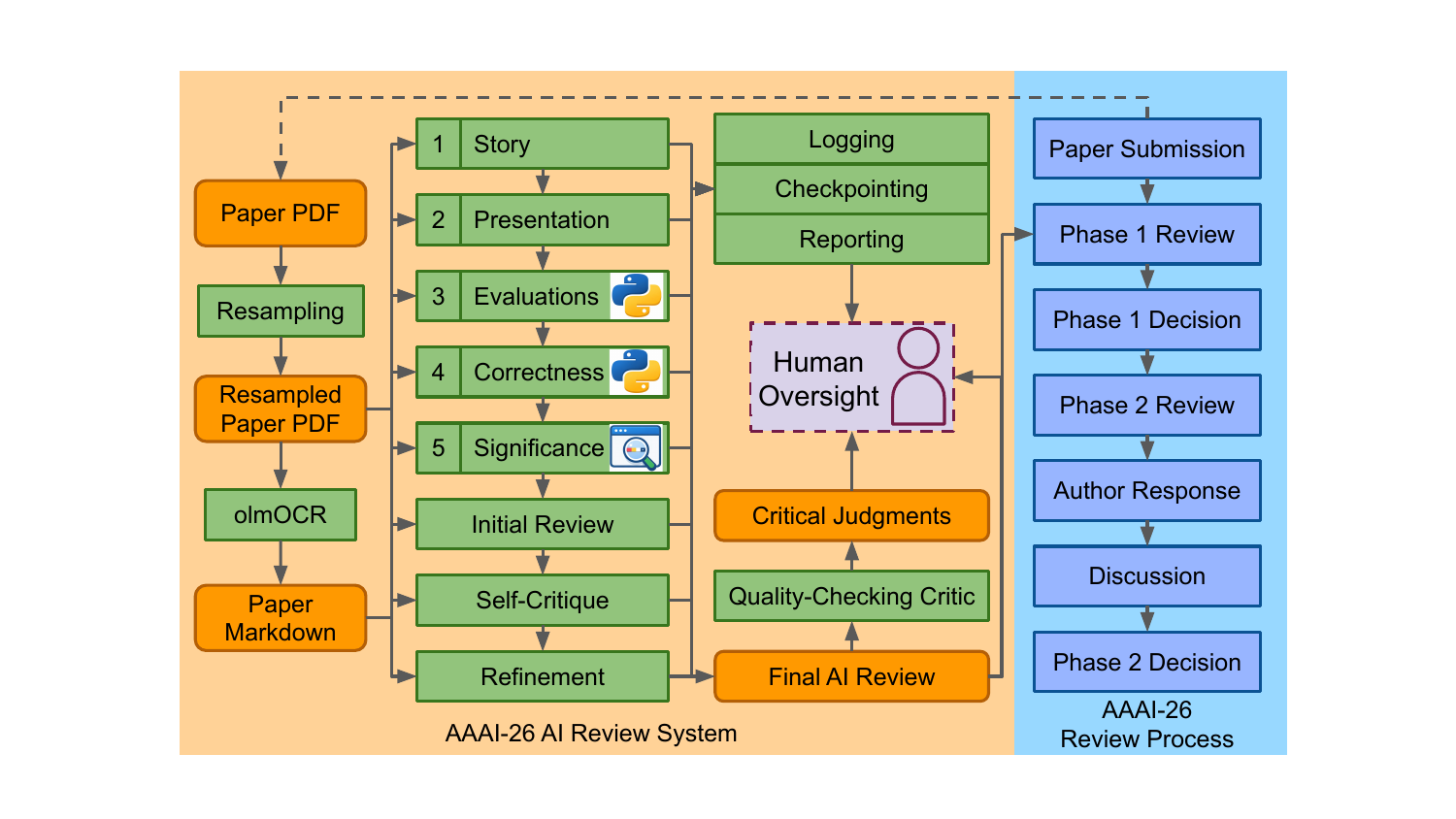}
        \caption{Review system}
        \label{fig:review-system-pipeline}
    \end{subfigure}
    \hspace{0.005\textwidth}
    \begin{subfigure}[c]{0.43\textwidth}
        \centering
        \includegraphics[width=\textwidth]{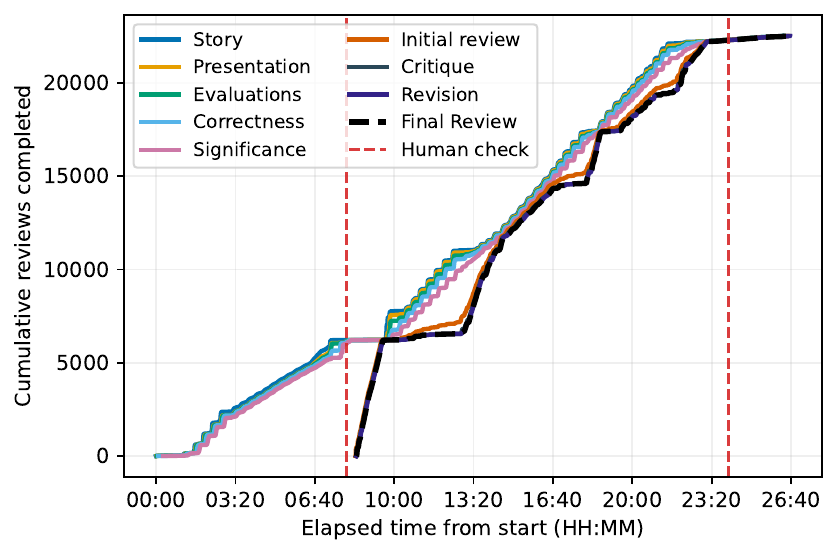}
        \caption{Review generation timeline}
        \label{fig:review-dashboard}
    \end{subfigure}
    \vspace{-0.4em}
    \caption{The AAAI-26 AI review system (a) and review generation timeline (b). For every submission to the AAAI-26 main track that entered the full review phase,
    the AI review system took as input the submitted paper in PDF form and generated an AI review that was then included in the first phase of the two-phase AAAI-26 review process. The AI review system first pre-processed the paper to resample all images to a consistent resolution of 250~DPI, and then used olmOCR~\cite{poznanski2025olmocr} to convert the PDF version of the paper to markdown. Both PDF and markdown versions of the paper were provided to the the subsequent stages. Five core review stages assessed the story, presentation, evaluations, correctness, and significance of the paper. An initial review was generated, compiling findings from all stages, into a consistent format. A self-critique stage checked the review for unsubstantiated claims, missing details, and inconsistencies with the paper. The final review stage revised the review to address the self-critique and compiled a final review. Logs, checkpoints, and review reports generated at all stages were saved for auditing and human oversight. The final review was assessed by a quality-checking critic, and a human inspected the critical judgments to identify potential concerns such as ethical concerns, revealing author identities, or missing structural elements in the review. The review generation timeline (b) shows the progress of all staged generations for all 22,977 papers. An initial batch of $30\%$ of the papers were run through the core review stages and the outputs inspected manually, before proceeding to complete all remaining stages and process all remaining papers. After the final manual inspection stage, a few papers were flagged for re-processing after manual PDF conversions to handle exceptional graphics formats. }
    \figlabel{review-system}
\end{figure*}

The AAAI-26 AI Review System integrates learnings from prior studies of AI-generated reviews~\cite{liang2024usefulfeedback,zhou-etal-2024-llm,zhang2025reviewingscientificpapers,aiscientist}.
A key design goal for the system was to ensure that the reviews considered scientific accuracy of all forms --- including mathematical and algorithmic correctness, sufficiency of the evaluation methods, and positioning of the work in the context of the previous state-of-the-art. 

Previous studies have shown that prompting LLMs with different `personas'~\cite{sahu2025reviewertoo,gao2025reviewagents,bougie2025gar} or criterion-specific prompts~\cite{liu2023reviewergpt}, rather than asking them to directly produce full reviews, can improve identification of specific types of scientific errors. 
Recent systems have also explored hierarchical question decomposition, deeper staged reasoning, and multimodal agent designs with shared memory for paper review~\cite{chang2025treereview,zhu2025deepreview,lu2025agentreviewers}.
The AAAI-26 AI Review System thus consists of five core scientific review stages intended to identify errors in: (1) story, (2) presentation, (3) evaluations, (4) correctness, and (5) significance.
The review system takes each PDF paper as input and generates
a textual review with markdown notation~\cite{gruber2012markdown} for math and tables, such that it can be rendered on the review interface.

\figref{review-system} shows the multi-stage AAAI-26 AI Review System that we built based on the desiderata and insights above.
After preprocessing, subsequent stages of the review system include both PDF and markdown versions of the paper, along with a system prompt to provide context for when to pay particular attention to one version over the other. Each review stage includes both a stage-specific prompt as well as the prompts and results from all previous stages. The evaluations and correctness stages include a Python code interpreter made available to the LLM to allow it to check for errors by testing out math and code snippets. The significance stage includes a web search tool to assist with literature search --- with specific instructions to restrict references to published work at relevant venues. After all targeted stages, the system generates an initial review and then revises it through a self-critique stage.
The initial review generation and revision prompts include specific instructions to ensure that each review contains the following structural elements: (1) the title of the paper, (2) a brief synopsis of the paper, (3) a summary of the review, (4) a detailed list of strengths, (5) a detailed list of weaknesses, and (6) a list of references cited in the review, in APA citation format. Prompt details for each stage are provided in \secref{appendix}.

We implemented a quality-checking workflow to identify potential issues in the generated reviews, similar to previous work on ``peer reviews of peer reviews''~\cite{goldberg2025peerreviewsofreviews}. Details on the quality-checking workflow and additional checks for citation hallucinations are provided in \secref{appendix}.

\section{Review Survey and Findings}
\seclabel{survey}
To assess the value and impact of the AI reviews on the AAAI-26 review process, we conducted a survey of all the participants: authors, PC, SPC, and ACs. Participation in the survey was voluntary, and the study design was reviewed by the University of Texas at Austin (IRB Protocol \texttt{STUDY00007931}), the University of Michigan (IRB Protocol \texttt{HUM00280758}), and the University of Alberta (REB Protocol \texttt{Pro00159777}).

\begin{table}[htb]
    \centering
    \small
        \caption{\small Number of responses by the respondent roles and review types, including the reviewer program committee (PC), senior program committee (SPC), and area chairs (AC). Response counts are further broken down by review type being surveyed (AI or human reviews). There were 5,834 responses in total.}
    \begin{tabular}{lrrrr}
        \toprule
        Review Type & Authors & PC & SPC & AC \\
        \midrule
        AI Reviews & 575 & 1305 & 117 & 12 \\
        Human Reviews & 2500 & 879 & 433 & 13 \\
        \midrule
        Total & 3075 & 2184 & 550 & 25 \\
        \bottomrule
    \end{tabular}

    \tablabel{survey-response-rate}
\end{table}

\begin{figure*}[t]
    \begin{minipage}[t]{0.56\linewidth}
        \vspace{0pt}
        \centering
        \subcaptionbox{AI \vs{} human review comparisons\figlabel{human-ai-differences}}{
            \includegraphics[width=\linewidth]{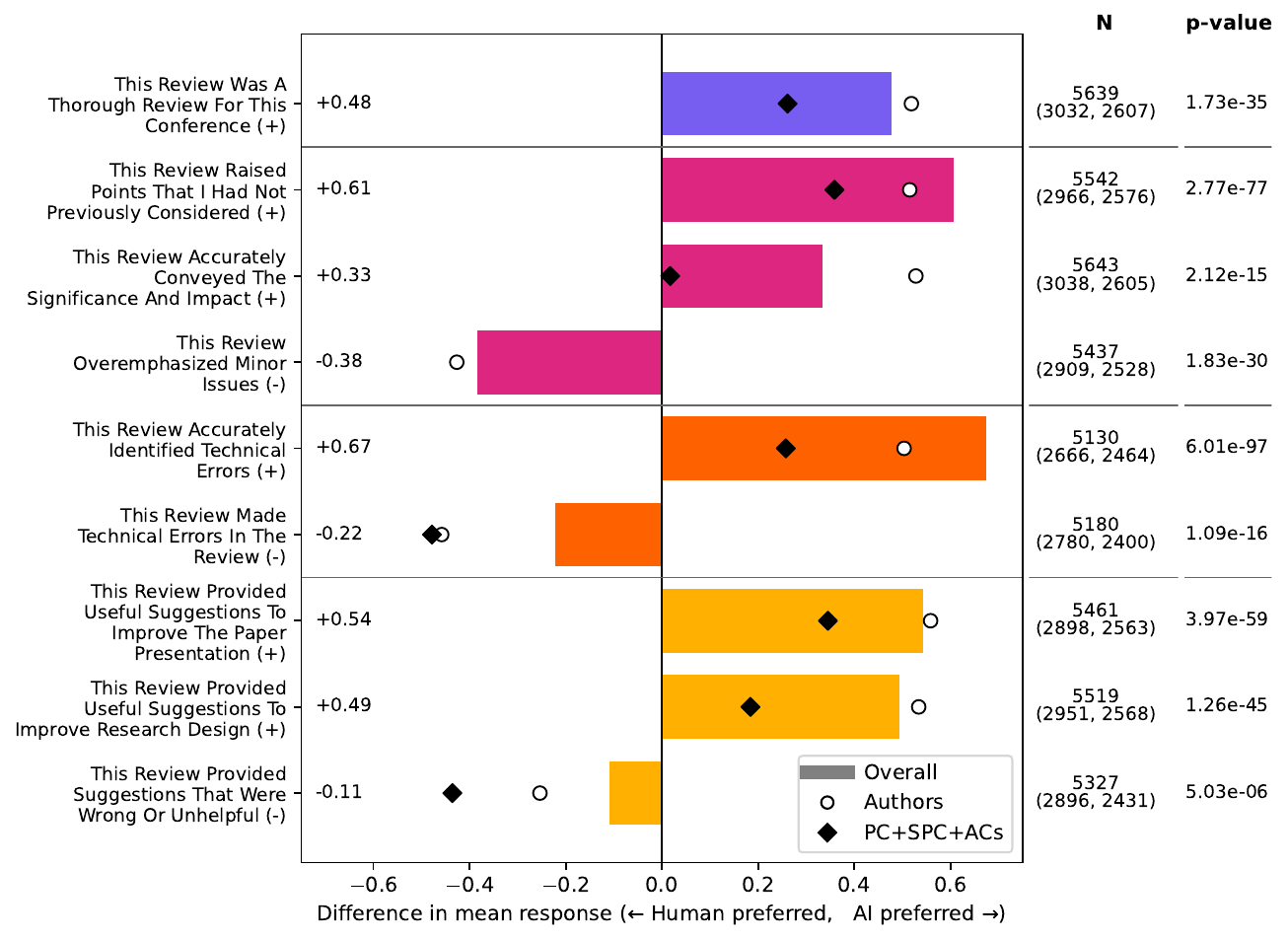}
        }
    \end{minipage}
    \begin{minipage}[t]{0.41\linewidth}
        \vspace{0pt}
        \centering
        \subcaptionbox{Responses to AI review questions\figlabel{ai-only-questions}}{
            \includegraphics[width=\linewidth]{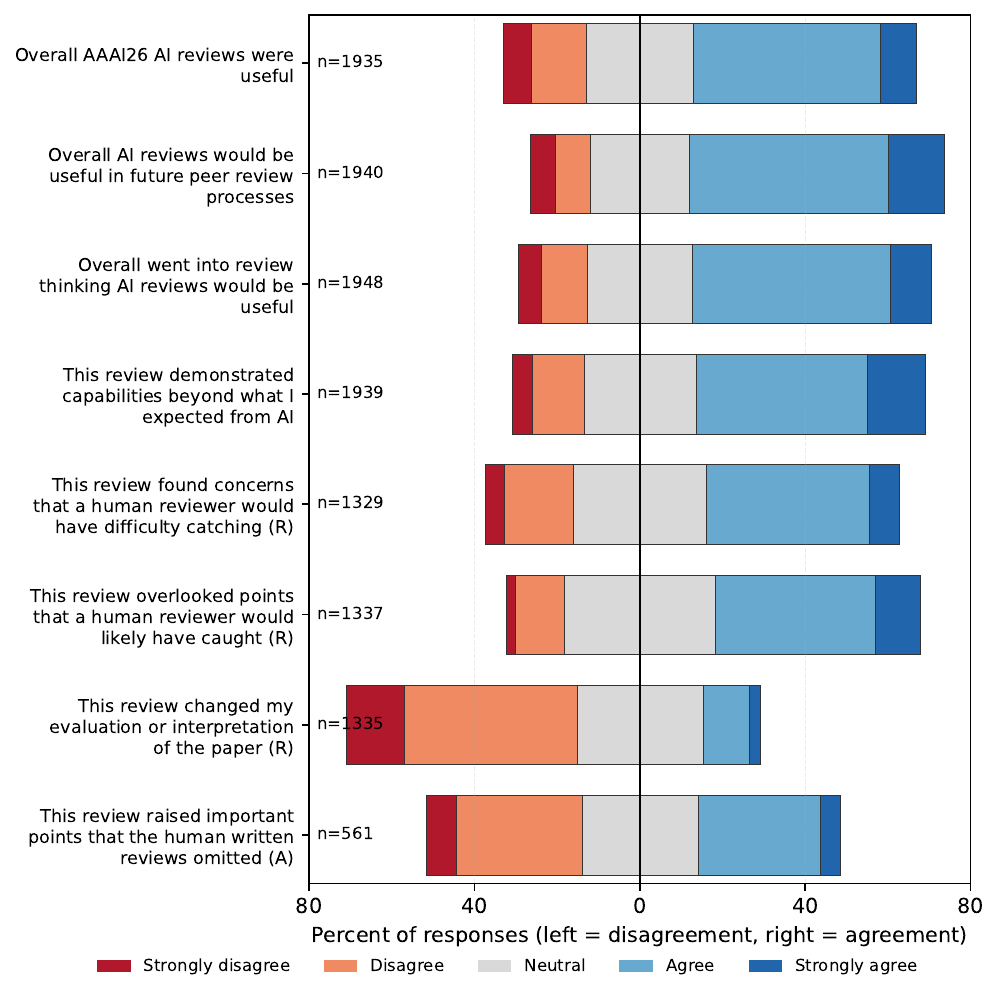}
        }
    \end{minipage}
    \caption{Survey responses: AI \vs{} human review comparisons (a) and AI review questions (b). The left figure shows the differences in the mean response score between AI and human reviews for each of the nine review-quality criteria. In six out of nine criteria, AI reviews were rated higher than human reviews. The preference towards AI reviews was stronger for authors than for PC, SPC, and ACs. All p-values show strong statistical significance at the $\alpha = 0.01$ level.
    The right figure shows the responses to the questions posed for just the AI reviews. Overall, there is strong support among the respondents that the AI reviews were useful in the AAAI-26 pilot and that they would be useful in future peer review processes. Respondents also expressed that the AI reviews demonstrated capabilities that were beyond what they expected from AI, despite indicating that they went into the process expecting that the AI reviews would be useful. Responses to questions posed specifically to the reviewers (labelled `(R)') show that the AI reviews both raised points that a human reviewer would have difficulty catching, while simultaneously overlooking points that a human reviewer would likely have caught --- indicating the complementary strengths of AI \vs{} human reviews. The majority of reviewers indicated that the AI reviews did not change their evaluation or interpretation of the paper, though a minority did indicate that they did. Authors (labelled `(A)') were mixed in their views about the AI reviews raising points that the human reviews omitted.}
    \figlabel{survey-results}
\end{figure*}

\begin{figure*}[htb]
    \centering
    \includegraphics[width=\linewidth]{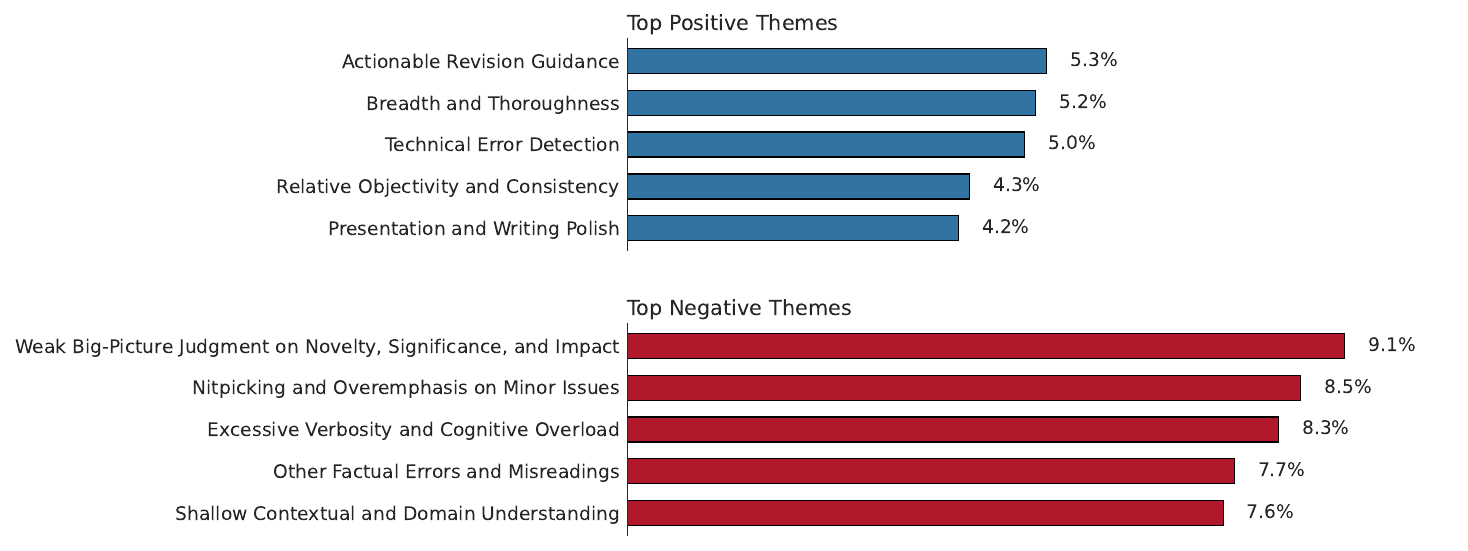}
    \caption{Top five most frequent positive and negative themes specific to the AAAI-26 AI Review Pilot found in written feedback from authors and program committee members. The percentages refer to the percentage of mentions belonging to that particular theme over all classified mentions specific to this pilot.}
    \label{fig:theme_freq}
\end{figure*}

\subsection{Survey Description}
The survey had a questionnaire associated with every review (both human and AI) that was made available to authors, PC, SPC, and ACs.  Authors and reviewers received the survey when the reviews were made available to them. For papers rejected in phase 1, this happened with the release of the phase 1 decisions, while for papers that proceeded to phase 2, this happened at the start of the author rebuttal period. 

Each questionnaire consisted of a set of statements about the review, with a five-point Likert scale to assess the degree of agreement with the statement (-2: strongly disagree, -1: disagree, 0: neutral, 1: agree, 2: strongly agree). The statements were grouped into four sections: (1) overall impressions, (2) review emphasis, (3) technical accuracy, and (4) suggestions for research and writing. An open-ended field allowed free-form comments to capture details about the review, the AAAI-26 AI Review Pilot, or any other comments related to the future of AI in peer review.

The survey received a total of 5,834 responses from authors, PC, SPC, and ACs, enabling strong statistical significance in the comparisons of AI and human reviews among respondents, though self-selection bias must also be considered. \tabref{survey-response-rate} shows the number of responses by role, for AI and human reviews.

\subsection{Quantitative Analysis}
We analyzed the survey responses to assess the overall attitudes toward the AAAI-26 AI Review Pilot, including comparing the distributions of responses for AI and human reviews. \figref{survey-results} shows the primary findings from the survey. The full survey text is included in \secref{appendix}.

\newcommand{\percent}[1]{\ensuremath{#1\%}}

We compared AI and human reviews on the nine review-quality criteria shown in \figref{human-ai-differences} by comparing the means of the empirical response distributions for each question. Overall, \textbf{AI reviews were preferred to human reviews on six of the nine criteria}, and all nine AI-human differences were statistically significant under a Mann-Whitney U test at $\alpha = 0.01$. The largest advantages for AI reviews were in identifying technical errors (+0.67), raising previously unconsidered points (+0.61), suggesting improvements to presentation (+0.54) and research design (+0.49), and overall thoroughness (+0.48). At the same time, respondents also judged AI reviews as more likely to overemphasize minor issues ($-0.38$), somewhat more likely to contain technical errors themselves ($-0.22$), and slightly more likely to include wrong or unhelpful suggestions ($-0.11$). The effect sizes were consistently larger for authors than for PC, SPC, and AC respondents.

We next consider the AI-only questions in \figref{ai-only-questions}. Overall, \percent{53.9} of respondents judged the AI reviews useful, compared with \percent{20.2} who did not, and \percent{61.5} expected AI reviews to be useful in future peer review, compared with \percent{14.5} who did not. Reported expectations before the pilot were already  positive, with \percent{57.8} expecting AI reviews to be useful and \percent{16.8} not expecting them to be useful. Even so, \percent{55.6} reported that the AI reviews demonstrated capabilities beyond what they had expected from AI, whereas \percent{17.3} did not.

The final set of questions addressed the AI reviews' complementary role alongside human reviews. Among PC, SPC, and AC respondents, \percent{46.6} agreed that the AI review found concerns that a human reviewer would have had difficulty catching, whereas \percent{21.3} disagreed. At the same time, \percent{49.4} agreed that the AI review overlooked points that a human reviewer would likely have caught, whereas \percent{14.0} disagreed, indicating that respondents saw AI reviews as complementary rather than interchangeable with human review. Consistent with this interpretation, only \percent{13.8} reported that the AI review changed their evaluation or interpretation of the paper, whereas \percent{55.6} disagreed. Authors were similarly mixed on whether the AI review raised important points omitted by the human reviews, with \percent{34.4} agreeing and \percent{37.6} disagreeing. 
The full distributions of responses for the overall thoroughness of human \vs{} AI reviews, as reported by authors and separated by whether the paper was accepted, are provided in \secref{appendix}. 
We found that among both accepted and rejected papers, AI reviews were rated as more thorough than human reviews, and the distributions for human reviews were wider than 
the distributions for AI reviews.

\subsection{Qualitative Analysis}
There were 320 usable free-form responses to the open-ended written feedback section in the survey.
Details on the process for aggregating a collective taxonomy of themes, which we then used to classify the points made in each free-form response, are provided in \secref{appendix}.

Fig.~\ref{fig:theme_freq} describes the five most frequent positive and the five most frequent negative themes specific to the AAAI-26 AI Review Pilot found in all the written feedback.  
Among the positive themes, the most frequently discussed was the aspect of \textit{Actionable Revision Guidance}, which describes how AI reviews tend to turn criticisms into actionable revision suggestions. Other aspects which were thought to be useful were the AI reviews' thorough and exhaustive analysis (\textit{Breadth and Thoroughness}), the flagging of technical mistakes overlooked by human reviewers (\textit{Technical Error Detection}), the consistent and objective nature of the reviews compared to humans (\textit{Relative Objectivity and Consistency}), and the reviews' detection of typos and other presentation errors (\textit{Presentation and Writing Polish}). In general, AI reviews appeared to be valued for their actionable and thorough feedback. 

\begin{figure*}[ht]
    \centering
    \centering
    \begin{subfigure}[c]{0.44\textwidth}
        \centering
        \includegraphics[width=\textwidth, trim=4.5cm 1.5cm 4.5cm 1.25cm, clip]{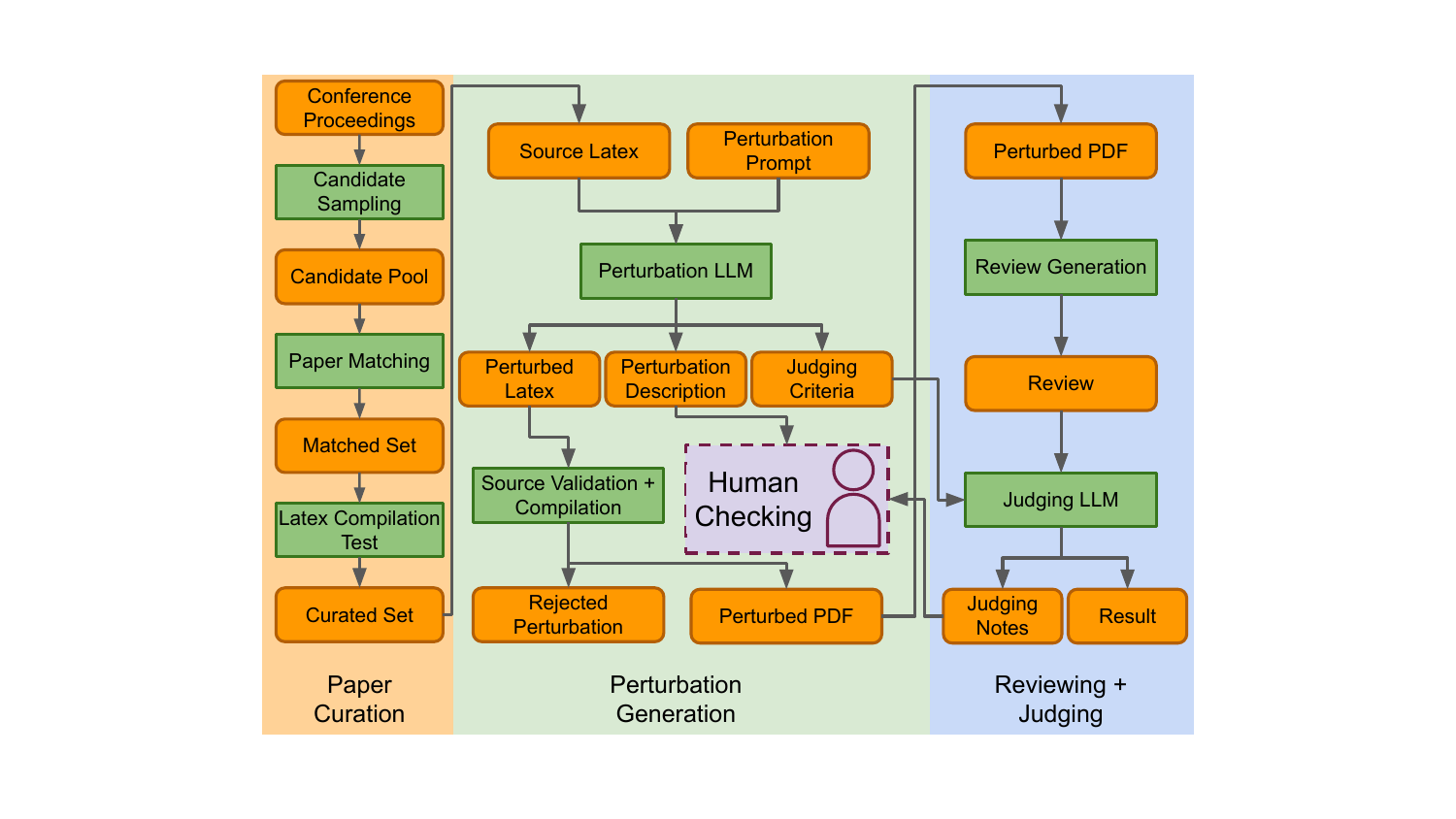}
        \caption{The SPECS benchmark}
        \label{fig:benchmark-curation-system}
    \end{subfigure}
    \hspace{0.005\textwidth}
    \begin{subfigure}[c]{0.53\textwidth}
        \centering
        \includegraphics[width=\textwidth, trim=0 0.125cm 0 0, clip]{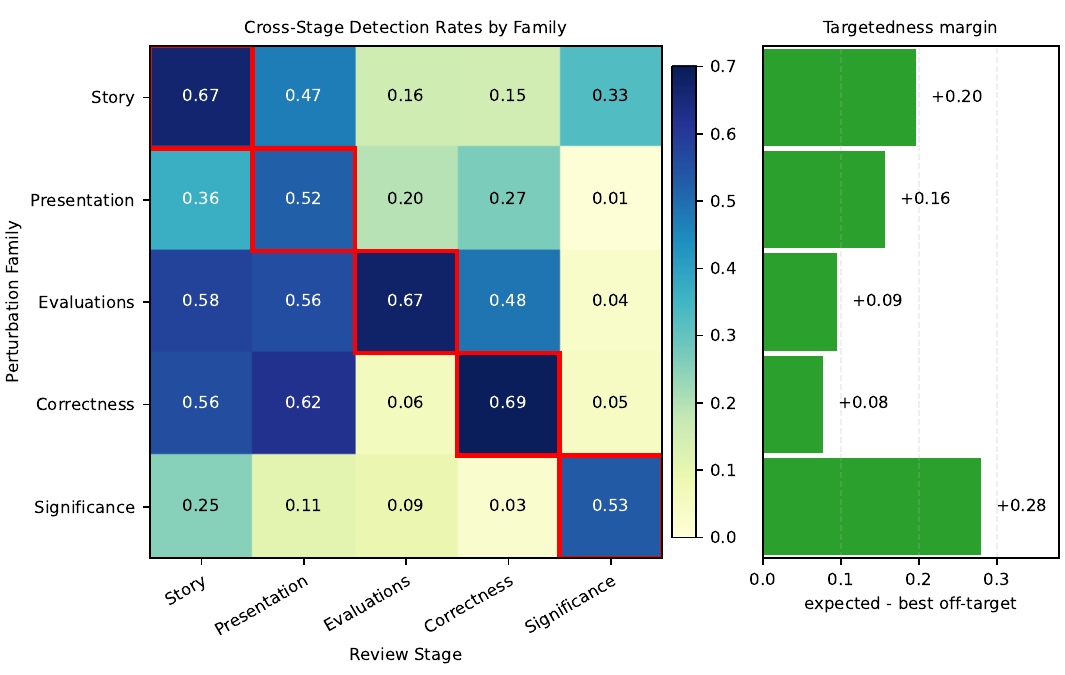}
        \caption{Stage-by-criterion detection rates}
        \label{fig:stage-confusion-matrix}
    \end{subfigure}
    \vspace{0.4em}
    \caption{The SPECS Review Benchmark curation and analysis workflow (a) and stage-by-criterion detection rates (b) from evaluating each stage in the AAAI-26 AI Review System for the SPECS criteria. Starting from the accepted papers listed in a venue proceedings (AAAI-25 in our case), the SPECS curation system samples across proceedings categories, matches each paper to an \arxiv{} source release, and retains papers whose \LaTeX{} sources compile successfully. For each curated paper, an LLM generates controlled source-level perturbations targeting the five SPECS criteria: \textbf{S}tory, \textbf{P}resentation, \textbf{E}valuations, \textbf{C}orrectness, and \textbf{S}ignificance. Each perturbation is accepted only if the cited source span matches the original file and the edited paper recompiles without errors, yielding executable synthetic errors for evaluation. The full AAAI-26 AI Review System and the criterion-targeted intermediate stages are then run on these perturbed papers, and a judge model determines whether each review explicitly identifies the injected error with supporting evidence from the review text. The stage-by-criterion detection-rate matrix in (b) summarizes criterion-wise detection behavior: each row corresponds to a perturbation type, while each column corresponds to a core stage in the AAAI-26 AI Review System. Each cell is computed independently as the fraction of perturbations of the row criterion detected by the column stage; rows do not sum to 1 because a perturbation can be detected by multiple stages (or none). The diagonal entries indicate criterion-aligned detection of the target core stage and the corresponding criterion, while off-diagonal entries reveal cross-criterion detection patterns. 
    The targetedness margin is the difference between a row's diagonal value and its largest off-diagonal value, indicating how much more strongly each stage detects its intended criterion than the other stages.}
    \figlabel{specs-curation}
\end{figure*}

The most frequently discussed negative point was that reviews often failed to accurately assess the novelty, significance of contributions, and overall scientific impact of a paper (\textit{Weak Big-Picture Judgment of Novelty, Significance, and Impact}). 
Other commonly discussed negative points concerned the AI review focus on minor issues (\textit{Nitpicking and Overemphasis on Minor Issues}), while overly lengthy AI reviews created additional cognitive work for authors and other reviewers (\textit{Excessive Verbosity and Cognitive Overload}). Such verbosity and nitpicking highlights the downside of the thoroughness found in AI reviews. 
Finally, the feedback also discussed \textit{Factual Errors and Misreadings} found in AI reviews, indicative of a continuing gap in LLMs understanding scientific content compared to an academic (human) established within their field.  
This ties into the the fifth most discussed theme (\textit{Shallow Contextual and Domain Understanding}), which specifically points out how the AI reviews struggled to provide feedback that was appropriate to a particular research domain. 

9 out of the 33 themes we found within the written feedback were general opinions about the use of AI in academic review process that were not specific to the AAAI-26 AI Review Pilot. Representative free-form survey responses illustrating these patterns are provided in \secref{appendix}.
Respondents also spoke positively about using AI as a pre-submission assistance tool and a means of generating meta-level summaries \cite{hossain-etal-2025-llms}. They highlighted its potential to scale peer review and support overburdened reviewers, while acknowledging that these systems are poised for rapid future improvement.
Not all of these opinionated themes were positive. Respondents also emphasized that AI reviews had the potential to mislead reviewers and other decision-makers in the review process.
There were also concerns that authors might optimize papers for AI preferences rather than scientific quality, and that reliance on these tools could lead to a long-term decline in reviewing skill. Adding to this, many respondents voiced principled objections, arguing that the use of AI undermines the trust, human effort, and essential value of the peer review process.

\section{The SPECS Review Benchmark}
\seclabel{benchmark}
A full peer review of a paper involves evaluating multiple criteria, including the 
(1) problem definition and formulation;
(2) technical accuracy of the proposed approach;
(3) sufficiency of the evaluation methods;
(4) accurate positioning of the work in the context of the related work; 
and (5) readability and clarity of the paper.
While there are several existing benchmarks for evaluating AI- and human-written reviews, they either focus on a specific criterion~\cite{xi2025flaws}, require specific structured outputs rather than full reviews~\cite{lou2025aaar,xi2025flaws}, or assess consistency of review scores between human and AI reviews~\cite{aiscientist,sahu2025reviewertoo,zhou-etal-2024-llm,zhang2025reviewingscientificpapers}.

\begin{table*}[htb]
    \centering
    \small
        \caption{SPECS benchmark results by criterion, comparing the AAAI-26 AI Review System (Final) to the baseline (Baseline) and criterion-targeted stages (Targeted). Improvements over baseline are listed in the $\Delta$ columns, and corresponding p-values are reported for each comparison.}
    \tablabel{specs-results}
    \begin{tabular}{lcccccccc}
    \toprule
    Criterion & $n$ & Baseline & Targeted & Final & $\Delta$ T--B & $p$-value & $\Delta$ F--B & $p$-value \\
    \midrule
    Story & 153 & 0.3529 & 0.6667 & 0.6732 & +0.3137 & $2.9\times 10^{-12}$ & +0.3203 & $1.5\times 10^{-12}$ \\
    Presentation & 173 & 0.4162 & 0.5202 & 0.5665 & +0.1040 & 0.0051 & +0.1503 & $4.2\times 10^{-5}$ \\
    Evaluations & 159 & 0.5157 & 0.6730 & 0.7547 & +0.1572 & 0.0006 & +0.2390 & $1.6\times 10^{-9}$ \\
    Correctness & 144 & 0.6111 & 0.6944 & 0.7639 & +0.0833 & 0.0576 & +0.1528 & $2.7\times 10^{-5}$ \\
    Significance & 154 & 0.2597 & 0.5325 & 0.4481 & +0.2727 & $6.5\times 10^{-7}$ & +0.1883 & 0.0003 \\
    \midrule
    All SPECS criteria & 783 & 0.4291 & 0.6143 & 0.6386 & +0.1852 & $4.8\times 10^{-21}$ & +0.2095 & $6.1\times 10^{-30}$ \\
    \bottomrule
    \end{tabular}

\end{table*}

We introduce the SPECS review benchmark to evaluate the effectiveness of AI review generation algorithms over multiple criteria, and to provide a comprehensive assessment of the quality of AI reviews. Specifically, it assesses the ability 
to catch errors in the
\textbf{S}tory,
\textbf{P}resentation,
\textbf{E}valuations,
\textbf{C}orrectness, and 
\textbf{S}ignificance of the paper.
The SPECS review benchmark follows a data curation and evaluation process similar to the FLAWS review benchmark~\cite{xi2025flaws}.
While FLAWS focuses on just technical errors and evaluates LLM responses in a structured output format, SPECS evaluates multiple criteria in a full free-form review format.
\figref{specs-curation} shows the process of curating the SPECS review benchmark dataset. The SPECS benchmark includes both a \emph{process} and a \emph{dataset} resulting from the process. The process can be repeated to generate a new dataset, \eg{} for a different venue, or to generate a new set of perturbations. 

Details on the curation process and the dataset that we generated by running the process on the AAAI-25 proceedings are provided in the Appendix. We focus here on the primary results from using the SPECS benchmark to evaluate the AAAI-26 AI Review System.

Given the set of PDFs compiled from the synthetic perturbations in the SPECS benchmark, the full AAAI-26 AI Review System was used to generate a review for each synthetically pertubed paper. Results from each review stage were additionally logged for analysis and comparison. As a baseline, we generated reviews for all papers using a single review prompt with no intermediate stages. 
This yielded a total of 5,481 reviews: 783 reviews from the baseline single-prompt approach, 3,915 reviews from the five intermediate review stages (one review per perturbation, per review stage), and 783 reviews from the final review stage. The reviews were judged (details in \secref{appendix}) to identify whether they caught the specific error introduced in the perturbation stage for the corresponding paper.

\tabref{specs-results} reports the recall rates of the AAAI-26 AI Review System (column `Final'), the baseline (column `Baseline'), and the criterion-targeted stages (column `Targeted') for each SPECS criterion. The `$\Delta$' columns show the absolute gains over baseline, and the `$p$-value' columns show the statistical significance of the gain compared to baseline, computed using a two-sided exact McNemar test. 
For every criterion, the AAAI-26 AI Review System significantly improves upon the baseline, with statistical significance at the $\alpha = 0.01$ level and an average gain of $+0.21$ across all criteria. Of the five targeted stages, all but the correctness stage demonstrate statistically significant improvements over the baseline, with an average gain of $+0.19$ across all criteria. Note that for one criterion, Significance, some errors are correctly identified in the targeted stage of the pipeline, but the final generated review does not explicitly identify them.
\figref{stage-confusion-matrix} shows the stage-by-criterion detection rates for the AAAI-26 AI Review System, revealing that each stage is indeed most effective (compared to other stages) at catching errors for its intended criterion.

\section{Conclusions}
The AAAI-26 AI Review Pilot Program demonstrated that AI-generated peer reviews are
operationally feasible at conference scale, and are capable of generating reviews that are helpful to the authors and reviewers.
The large-scale survey of AAAI-26 authors, reviewers, senior program committee members, and area chairs found that participants broadly found AI reviews useful and preferred them to human reviews on key dimensions such as technical accuracy and research suggestions, but also identified some limitations and areas for improvement including technical errors in reading some equations and tables, difficulty in prioritizing the significance of issues, and producing reviews that were longer than readers preferred. The quantitative and qualitative analyses combined indicate the complementary strengths of AI systems and human reviewers, and suggest that future work should explore how best to integrate the two to leverage their strengths.

\section{Acknowledgments}
We thank the AAAI Executive Council, Ethics Committee, and Conference Committee for their support and guidance in shaping the AAAI-26 AI Review Pilot Program.
Zico Kolter initiated the idea of the AI review pilot, and served as liaison to OpenAI for securing a sponsorship to support the initiative.
We thank OpenAI for providing API credits as a AAAI-26 sponsor to support the development, deployment, and evaluation of the AI review system.
Meredith Ellison (AAAI Executive Director) and Marc Pujol (AAAI Director of Program Operations \& Systems) assisted in the operations and logistics of deploying the AI review pilot for AAAI-26.
We are grateful to the OpenReview staff for their assistance in integrating the AAAI-26 AI Review System reviews and survey forms into the OpenReview platform.

\let\bibliographystyle\aaaiorigbibliographystyle
\clearpage
\bibliographystyle{unsrtnat}
\bibliography{references}

\clearpage
\appendix
\section{Appendix}
\seclabel{appendix}

\subsection{Prompt design}
We report the prompt structure and design for the AAAI-26 AI Review System. The exact prompts are withheld to reduce the risk of prompt-targeted optimization or prompt injection attacks in future deployments.

The review pipeline uses a layered prompt design rather than a single monolithic instruction. This sequence of decomposition, synthesis, and validation is intended to improve coverage across all review criteria and minimize errors in the review process.
A persistent base instruction establishes the overall reviewing objective, the expectation of factual and objective analysis, and guidance on how to use the PDF and OCR-derived markdown versions of the paper together. This base instruction is reused across all later stages so that each stage operates with the same review context and source-handling policy.

On top of this shared base layer, the system uses a fixed stack of specialized prompt components, listed in \tabref{prompt-inventory}.
These components are invoked in a defined order and play distinct roles in the pipeline: a base review instruction, the story, presentation, evaluations, correctness, and significance prompts, an initial review prompt, a self-critique prompt, and a final review prompt. The main purpose of this decomposition is to reduce overload, encourage deeper checking of each review criterion, and preserve intermediate findings for later synthesis. 

Upon completion of all core stages, the system uses a sequence of prompts to compile the findings into a single review, generate a self-critique, and revise the review to address the self-critique to produce the final review. 

\begin{table*}[t]
\centering
\small
\caption{Descriptions for prompts used in the AAAI-26 AI review system. The table lists the prompt components, where each is used, and a high-level synopsis of what each prompt asks the model to do.}
\begin{tabular}{p{0.20\textwidth}p{0.21\textwidth}p{0.5\textwidth}}
\toprule
\textbf{Prompt component} & \textbf{Where it is used} & \textbf{Synopsis of Prompt} \\
\midrule
Base instruction & Prepended to all stages & Explains the task to the model: to generate a factual, objective review without scores or recommendations through a sequence of subsequent steps.  Also explains how to use the PDF and OCR-derived markdown as complementary sources. \\
Story prompt & Story stage & Asks the model to analyze the paper's problem formulation, claimed gap in prior work, core contribution, and whether the evidence supports the main story. \\
Presentation prompt & Presentation stage & Asks the model to assess clarity, organization, readability, and whether the technical narrative is easy for a researcher to follow. \\
Evaluations prompt & Evaluations stage & Asks the model to inspect baselines, datasets, metrics, statistical evidence, empirical support for claims, and reproducibility-related weaknesses, using the available code interpreter tool if needed. \\
Correctness prompt & Correctness stage & Asks the model to verify equations, proofs, algorithms, figures, and tables, using the available code interpreter tool if needed. \\
Significance prompt & Significance stage & Asks the model to identify closely related published work at top-tier venues and use it to contextualize novelty, competitiveness, and missing comparisons, with specific instructions to restrict results to published work only, and to disregard preprints or non-peer-reviewed work. \\
Initial review prompt & Initial review stage & Asks the model to synthesize the accumulated findings into a standardized draft review with a fixed structure. Includes details about the specific markdown variant used by the frontend, and formatting instructions to ensure rich text formatting including math notation and tables, if any, are rendered correctly. \\
Self-critique prompt & Self-critique stage & Asks the model to reread the draft review against the paper and flag unsupported claims, factual problems, and unclear or incomplete citations. \\
Final review prompt & Final review stage & Asks the model to revise the draft to address the self-critique and compile a final review. \\
\bottomrule
\end{tabular}

\tablabel{prompt-inventory}
\end{table*}

\subsection{Quantitative Analysis Details}
For each question that was included in both the AI and human review questionnaires, we compared the responses for AI and human reviews.
Let $R^\mathrm{human}_\mathrm{authors}$ and $R^\mathrm{AI}_\mathrm{authors}$ be the collections of responses for the human and AI review questionnaires from the authors, respectively,
and let $R^\mathrm{human}_\mathrm{reviewers}$ and $R^\mathrm{AI}_\mathrm{reviewers}$ be the collections of responses for the human and AI reviewers questionnaires, respectively.
Let $M(R)$ be the mean of the response collection $R$:
\begin{equation*}
    M(R) = \frac{1}{|R|} \sum_{i=1}^{|R|} R_i,
\end{equation*}
where $|R|$ is the number of responses in the collection $R$.
The difference in the distribution means for the human and AI reviews for the authors and reviewers separately are given by
\begin{equation*}
    M(R^\mathrm{AI}_\mathrm{authors}) - M(R^\mathrm{human}_\mathrm{authors}),
\end{equation*}
and
\begin{equation*}
    M(R^\mathrm{AI}_\mathrm{reviewers}) - M(R^\mathrm{human}_\mathrm{reviewers}),
\end{equation*}
respectively.
The overall mean $M(R^\mathrm{AI}_\mathrm{overall})$ for the AI reviews from all respondents (both authors and reviewers) is given by
\begin{equation*}
    \frac{|R^\mathrm{AI}_\mathrm{authors}| M(R^\mathrm{AI}_\mathrm{authors}) + |R^\mathrm{AI}_\mathrm{reviewers}| M(R^\mathrm{AI}_\mathrm{reviewers})}{|R^\mathrm{AI}_\mathrm{authors}| + |R^\mathrm{AI}_\mathrm{reviewers}|},
\end{equation*}
and the overall mean $M(R^\mathrm{human}_\mathrm{overall})$ for the human reviews from all respondents (both authors and reviewers) is given by
\begin{equation*}
    \frac{|R^\mathrm{human}_\mathrm{authors}| M(R^\mathrm{human}_\mathrm{authors}) + |R^\mathrm{human}_\mathrm{reviewers}| M(R^\mathrm{human}_\mathrm{reviewers})}{|R^\mathrm{human}_\mathrm{authors}| + |R^\mathrm{human}_\mathrm{reviewers}|}.
\end{equation*}
The overall difference in the distribution means for the human and AI reviews is thus given by
\begin{equation*}
    M(R^\mathrm{AI}_\mathrm{overall}) - M(R^\mathrm{human}_\mathrm{overall}),
\end{equation*}
We used a Mann-Whitney U test to assess the statistical significance of the differences in the response distributions for AI and human reviews, testing against the null hypothesis that the samples are drawn from the same distribution. For example, for the author responses, we tested the null hypothesis that $R^\mathrm{AI}_\mathrm{authors}$ and $R^\mathrm{human}_\mathrm{authors}$ are drawn from the same distribution, against the alternative hypothesis that the two response collections are drawn from different distributions. This test was performed for each AI \vs{} human review comparison for overall, authors, and reviewers separately.

\subsection{Survey Questionnaire}
Table~\ref{tab:survey-questionnaire} summarizes the questionnaire items shown to respondents. There were four questionnaire variants, corresponding to whether the respondent was an author or a reviewer (PC, SPC, or AC), and whether the questionnaire was attached to an AI review or a human review. All closed-form items used the same five-point Likert scale described in \secref{survey}, and each questionnaire included an open-ended field for free-form comments.

\begin{table*}[t]
\centering
\caption{Questionnaire items by respondent group and review type. `A' denotes author questionnaires and `R' denotes the shared reviewer-form structure used for PC, SPC, and AC respondents. `AI' and `H' denote questionnaires attached to AI reviews and human reviews, respectively.}
\tablabel{survey-questionnaire}
\footnotesize
\setlength{\tabcolsep}{4pt}
\begin{tabular}{@{}p{0.70\textwidth}cccc@{}}
\toprule
\textbf{Questionnaire item} & \textbf{A/AI} & \textbf{A/H} & \textbf{R/AI} & \textbf{R/H} \\
\midrule
\multicolumn{5}{@{}l}{\textit{Overall impressions}} \\
This review was a thorough review for this conference & $\checkmark$ & $\checkmark$ & $\checkmark$ & $\checkmark$ \\
This review demonstrated capabilities beyond what I expected from AI & $\checkmark$ &  & $\checkmark$ &  \\
This review changed my evaluation or interpretation of the paper &  &  & $\checkmark$ &  \\
This review changed my evaluation of the paper &  &  &  & $\checkmark$ \\
Overall, AAAI-26 AI reviews were useful & $\checkmark$ &  & $\checkmark$ &  \\
Overall, AI reviews were harmful in AAAI-26 & $\checkmark$ &  & $\checkmark$ &  \\
Overall, AI reviews would be useful in future peer review processes & $\checkmark$ &  & $\checkmark$ &  \\
Overall, AI reviews would be harmful in future peer review processes & $\checkmark$ &  & $\checkmark$ &  \\
Overall, I went into review thinking AI reviews would be useful & $\checkmark$ &  & $\checkmark$ &  \\
\midrule
\multicolumn{5}{@{}l}{\textit{Review emphasis}} \\
This review accurately conveyed the significance and impact & $\checkmark$ & $\checkmark$ & $\checkmark$ & $\checkmark$ \\
This review overemphasized minor issues & $\checkmark$ & $\checkmark$ & $\checkmark$ & $\checkmark$ \\
This review raised points that I had not previously considered & $\checkmark$ & $\checkmark$ & $\checkmark$ & $\checkmark$ \\
This review raised important points that others might have missed &  & $\checkmark$ &  & $\checkmark$ \\
This review raised important points that the human-written reviews omitted & $\checkmark$ &  &  &  \\
This review found concerns that a human reviewer would have difficulty catching &  &  & $\checkmark$ &  \\
This review overlooked points that a human reviewer would likely have caught &  &  & $\checkmark$ &  \\
This review discovered important points that AI would have missed &  &  &  & $\checkmark$ \\
This review overlooked points that AI would have caught &  &  &  & $\checkmark$ \\
\midrule
\multicolumn{5}{@{}l}{\textit{Technical accuracy}} \\
This review accurately identified technical errors & $\checkmark$ & $\checkmark$ & $\checkmark$ & $\checkmark$ \\
This review made technical errors in the review & $\checkmark$ & $\checkmark$ & $\checkmark$ & $\checkmark$ \\
\midrule
\multicolumn{5}{@{}l}{\textit{Suggestions for research and writing}} \\
This review provided useful suggestions to improve research design & $\checkmark$ & $\checkmark$ & $\checkmark$ & $\checkmark$ \\
This review provided useful suggestions to improve the paper presentation & $\checkmark$ & $\checkmark$ & $\checkmark$ & $\checkmark$ \\
This review provided suggestions that were wrong or unhelpful & $\checkmark$ & $\checkmark$ & $\checkmark$ & $\checkmark$ \\
This review correctly suggested related work & $\checkmark$ & $\checkmark$ & $\checkmark$ & $\checkmark$ \\
This review incorrectly suggested nonexistent work & $\checkmark$ &  & $\checkmark$ &  \\
This review raised points to be addressed in camera-ready or resubmission & $\checkmark$ & $\checkmark$ &  &  \\
This review raised points to be addressed in a revision &  &  &  & $\checkmark$ \\
\midrule
\multicolumn{5}{@{}l}{\textit{Free response}} \\
General feedback on the review and the AAAI-26 AI Review Pilot & $\checkmark$ & $\checkmark$ & $\checkmark$ & $\checkmark$ \\
\bottomrule
\end{tabular}

\end{table*}

\subsection{Preprocessing PDFs}
To ensure that the multimodal tokenization of the paper PDF does not exceed the context window limit of the LLM due to the inclusion of high-resolution images, all paper PDFs are resampled to a consistent resolution of 250~DPI.
We found through initial testing that the PDF version of the paper alone is insufficient for accurately reading equations and tables --- initial testing demonstrated errors in the generated reviews caused by the LLM misinterpreting mathematical notation and table structures. To overcome this issue, a specialized optical character recognition (OCR) model, olmOCR~\cite{poznanski2025olmocr}, is used to convert the PDF version of the paper to markdown. The markdown version includes \LaTeX{} notation for the equations and inline math symbols, and a structured layout for the tables in the paper.

\subsection{LLM Details for AAAI-26 AI Review System}
We used the OpenAI \texttt{gpt-5} LLM~\cite{singh2025openai} with `high' reasoning effort for all stages of the AAAI-26 AI Review System. This model was the most capable model among the OpenAI models available at the time of deployment, and had a September 30, 2024 knowledge cutoff.
All OpenAI API calls were made using a special Zero Data Retention (ZDR) agreement in place to ensure that confidentiality was upheld to the highest degree --- model inputs and outputs were not logged, and only existed in ephemeral memory on the OpenAI servers.
The context window for this model was 400,000 tokens with 128,000 max output tokens, which included the reasoning tokens in addition to the final output tokens. For PDF inputs, OpenAI's file-input documentation~\cite{openai2026fileinputs} states that vision-capable multimodal models process both extracted text and rendered page images, so PDF context usage reflects both textual and visual content rather than text alone.
The evaluations and correctness stages included the built-in \texttt{code\_interpreter} tool with `auto' container type.
The significance stage included the built-in \texttt{web\_search\_preview} tool with `approximate' user location set to `US' and `medium' search context size. All OpenAI LLM calls were made with the \texttt{flex} tier, with error-handling and retry logic implemented to ensure robustness.
During deployment, since review generation is executed in parallel for a large batch of papers, occasional service errors due to rate limits, server load, or other issues are inevitable.
Upon error, the LLM calls were retried up to 5 times with exponential backoff.

\subsection{Quality-Checking and Human Oversight of AI Reviews}
After all reviews were generated, a second critic LLM (OpenAI \texttt{o4-mini}) was prompted to identify specific potential issues in the reviews as well as editorial concerns about the papers. The critic LLM was not provided with the paper or any of the original prompts, and was not told that the reviews were generated by an AI system.

The list of review issues assessed included (1) revealing author identities; (2) potentially offensive language or content in the review; (3) judgments that may have been biased based on gender, geography, or other factors; and (4) missing structural elements in the review.
Editorial concerns included (1) ethical concerns, (2) inclusion of information in the paper that may reveal author identities, and (3) paper violations of conference policies such as formatting requirements.

We additionally asked the critic LLM to judge auxiliary criteria including (1) whether the review appeared to be written by an LLM, (2) whether the review appeared to be written by an unqualified reviewer, (3) the apparent effort that went into the review, and (4) an overall rating of the quality of the review.

The critic LLM's judgments were compiled into a spreadsheet, after which the authors manually inspected all reviews that were identified as potentially containing one or more of the above issues. Through this process, we identified several papers that violated author anonymity through one of several means including the inclusion of specific acknowledgments or links to public non-anonymous websites.
There were a few papers flagged for ethical concerns including potential software license violations and potential identity leakage in the dataset released with a paper.

\subsubsection{Citation Hallucination Checking}
We randomly sampled 100 reviews generated by the AAAI-26 AI review system and checked for hallucinated citations using the GPTZero~\cite{gptzero} API.
There were 1356 citations in the sampled reviews.
GPTZero identified 1346 of the citations as valid, and matched them to published work at the cited venues, with matching authors and titles.
It labelled 8 citations as `unsure,' and 2 citations as `fake.' We further manually inspected the 8 `unsure' citations and the 2 `fake' citations and found that the ones marked as `unsure' were indeed correctly matched to published work, and one of the `fake' citations actually existed, but was a (correct) citation to a technical reference manual rather than a published work, which the tool is designed to detect. The second `fake' citation had an incorrect venue --- the paper existed by the cited authors and paper title, but at a different venue than the one cited.

\subsection{Survey Taxonomy Creation and Feedback Classification}
We used the OpenAI \texttt{gpt-5.4} LLM with `high' reasoning effort to summarize the free-form survey responses into positive and negative categories, similar to previous LLM-assisted survey coding approaches~\citep{Parker2025,doi:10.1177/23328584251374595}.
The LLM was prompted to detail the name of the category, a one sentence summary of the category, and a more detailed description and rationale for why this category was created.
We initially extracted 13 positive themes and 20 negative themes through this process; the larger number of negative themes is consistent with the well-documented negativity bias in open-ended survey feedback, where dissatisfied respondents are more likely to comment and comments tend to be disproportionately negative in tone~\cite{rozin2001negativitybias,poncheri2008negativitybias}.
Of these categories, via manual inspection, we identified 24 out of the 33 categories as being specific to the AAAI-26 AI Review Pilot, \vs{} 9 as being general opinions about the use of AI in the academic review process, both present and future.

After creating this taxonomy, we applied it to extract individual instances of such categories mentioned in every piece of written feedback.
As in the taxonomy creation process,
we used \texttt{gpt-5.4} to analyze the written feedback and identify if any of the categories were mentioned in the feedback, and if so, to provide attribution for which excerpts in the feedback support which categories of feedback. Each feedback item could support multiple categories, and the LLM was prompted to provide a complete list of identified categories, with attribution.
The LLM was also prompted to provide a rationale for every category identified.

\subsection{Sample Free-Form Survey Responses}
Here, we show a couple of examples of the free-form survey responses we received that cover many of the commonly found themes, both positive and negative, among all responses.

\subsubsection{Anonymous Author Response 1}
\begin{quote}
    Overall, the AI review pilot demonstrates promising potential in enhancing the peer review process. Its most notable positive aspect is the unexpected thoroughness in technical scrutiny unlike some human reviews that might overlook niche details, the AI effectively identified overlooked technical gaps and provided contextually relevant, authentic references, which significantly aids authors in refining their work. Objectivity is another strength, as the AI's evaluations appear less influenced by subjective biases, focusing more on alignment with established methodologies and literature. However, a potential area for improvement lies in nuanced contextual understanding; while technically precise, the AI sometimes lacks the depth of insight that comes from a human expert s long-term immersion in a specific subfield, such as recognizing the broader, unstated implications of a study's limitations or the novelty of work that deviates slightly from mainstream approaches. Looking ahead, AI could best serve peer review as a complementary tool handling the initial screening of technical accuracy, reference verification, and structural coherence to free up human reviewers to focus on higher-level assessments of innovation, significance, and real-world impact, creating a more efficient and balanced process.
\end{quote}

\subsubsection{Anonymous Author Response 2}
\begin{quote}
    I appreciate being able to participate in the AI review pilot. This particular review felt impressively thorough, constructive, and fair---it engaged deeply with the technical and experimental aspects of the paper and made thoughtful and highly actionable suggestions regarding methodology, reproducibility, related work, and presentation. The level of detail and breadth of recent references exceeded my initial expectations for automated reviewing.

A notable strength is the AI's ability to check consistency, pinpoint technical ambiguities (such as token-retention accounting and KV-cache handling), and call for more rigorous baseline and efficiency comparisons. These are aspects that even experienced human reviewers sometimes miss or only mention superficially. The AI provided a rich set of pointers for both technical and presentational improvement.

On the other hand, while the AI review raised almost all the necessary concerns, it sometimes leaned toward exhaustive checklists rather than contextual prioritization or nuanced judgment about what issues most affect the overall scientific contribution. I also noticed that some points (e.g., writing/formatting inconsistency) might be less helpful in earlier-stage submissions where major revisions are expected.

Overall, I see significant potential for AI to support or supplement peer review, ensuring greater thoroughness and standardization. However, human insight, especially in evaluating novelty, potential real-world impact, and the balance between theoretical elegance and practical feasibility, remains indispensable. Combining AI's systematic coverage with human expertise could improve review quality and fairness.

Thank you for running this pilot---I found it thought-provoking and valuable.
\end{quote}

\subsection{SPECS Benchmark Curation Process}
We describe the SPECS benchmark curation process in detail. The LLM used for this process was OpenAI \texttt{gpt-5.4} with `high' reasoning effort.

\subsubsection{Paper Selection}
The benchmark dataset curation starts with a collection of recently accepted papers from a relevant venue --- AAAI-25 in our case.
It then samples papers to cover a representative set of topics across the proceedings category organization to yield a candidate pool of papers. For each paper, it searches for a matching paper on \arxiv{}, and downloads the \latex{} source code of the paper. A paper is included in the benchmark if (1) the bibliographic information in the proceedings matches that on \arxiv{}, and (2) the \LaTeX{} source code compiles successfully without errors. The curation process yielded 120 papers from the AAAI-25 proceedings for the benchmark, with a track distribution of Game Theory and Economic Paradigms (39, 32.5\%), Machine Learning (30, 25.0\%), Knowledge Representation and Reasoning (14, 11.7\%), Computer Vision (12, 10.0\%), Constraint Satisfaction and Optimization (11, 9.2\%), Data Mining \& Knowledge Management (8, 6.7\%), Intelligent Robots (4, 3.3\%), and Humans and AI (2, 1.7\%). The specific distribution was a result of the distribution of accepted papers in the AAAI-25 proceedings, random sampling from the distribution, the failure rate to find \arxiv{} source, and the failure rate to compile the \LaTeX{} source.

\subsubsection{Perturbation Generation}
For each paper, an LLM is prompted to generate synthetic edits to the \LaTeX{} source of the paper to introduce one of five perturbation types of scientific errors: story, presentation, evaluation, correctness, and significance. The prompt used for each perturbation type describes the perturbation type and includes specific examples. Perturbation types evaluation, correctness, and presentation include several subtypes to further classify the type of error --- for example, the evaluation subtypes include missing baseline, missing metric, and data misinterpretation. The LLM is prompted with (1) the perturbation type and subtype with their associated descriptions and examples and (2) the \LaTeX{} source code of the paper; and instructed to output (1) a description of the perturbation being introduced and why it is significant to catch in a peer review, (2) the original source \LaTeX{} being edited and the associated file name and line numbers, and (3) the modified \LaTeX{} source code.
A generated perturbation is accepted if the identified original source being edited matches the paper source at the correct file name and line numbers, and if the modified source compiles successfully without errors.
The perturbation process yielded 783 synthetic perturbations,
with a distribution of 153 (19.5\%) for story, 173 (22.1\%) for presentation, 159 (20.3\%) for evaluation, 144 (18.4\%) for correctness, and 154 (19.7\%) for significance.

\subsubsection{Judging}
Each review was judged by an LLM prompted to identify if the review caught the specific error introduced in the perturbation stage for the corresponding paper.
A review is considered to have caught the error only if it (1) explicitly identified the specific error, and (2) substantiated the identification with a text excerpt from the review that matches the synthetically perturbed text. To allow human audit of the judging results, the judge LLM was also prompted to output a justification for its judgment, including which excerpt from the review supported the judgment, or why the review did not catch the error.

\subsubsection{Human Oversight}
To ensure that the perturbations generated by the LLM were indeed significant scientific errors that should be identified in a peer review, a randomly sampled set of 35 perturbations were independently inspected by two human reviewers --- 6 story, 7 presentation, 8 evaluation, 7 correctness, and 7 significance perturbations.

\begin{table}[!htbp]
    \centering
    \small
    \caption{SPECS benchmark human oversight. Reviewers R1 and R2 independently reviewed each perturbation. The responses here are the verdicts of the perturbation that they declared valid.}
    \tablabel{specs-human-oversight}
    \begin{tabular}{lcccc}
    \toprule
    Criterion & $n$ & R1 & R2 & Consensus \\
    \midrule
    Story & 6 & 5 & 5 & 5 \\
    Presentation & 7 & 3 & 4 & 3 \\
    Evaluations & 8 & 5 & 6 & 5 \\
    Correctness & 7 & 7 & 6 & 6 \\
    Significance & 7 & 3 & 4 & 3 \\
    \midrule
    All SPECS criteria & 35 & 23 & 25 & 22 \\
    \bottomrule
    \end{tabular}
\end{table}

From the manual inspection, both reviewers agreed that 22 perturbations were valid scientific errors that should be identified in a peer review, and 9 were minor errors that did not warrant being noted in a peer review. The reviewers were split on 4 perturbations.
The presentations and significance perturbations were particularly difficult --- the reviewers unanimously assessed that only $3/7$ for both of these types were significant enough to make a significant impact on paper quality. The story, correctness, and evaluations perturbations, on the other hand, were relatively successful, with unanimous agreement on $5/6$, $6/7$, and $5/8$ respectively. The Appendix summarizes the human-oversight results by criterion (\tabref{specs-human-oversight}).
Of the valid perturbations, two additional reviewers reviewed 40 judging results from the LLM, randomly drawing from judging results from the baseline, targeted stages, and final system. All but one of the 40 judgments were unanimously agreed upon as being correct.

\end{document}